\documentclass[letterpaper, 10 pt, conference]{ieeeconf}
\IEEEoverridecommandlockouts
\usepackage{amsmath,amssymb, graphicx, epstopdf,cite,enumerate,booktabs, setspace, float,stfloats,bm, multirow, lscape}

\usepackage[normalem]{ulem}

\allowdisplaybreaks[4]
\usepackage{caption}
\usepackage{subcaption}
\usepackage{graphbox}
\usepackage{soul, color, xcolor}
\usepackage[bottom=1.52cm,top=2.01cm,left=1.69cm,right=1.69cm]{geometry}

\usepackage[linesnumbered,ruled]{algorithm2e}

\title{\LARGE \bf iCOIL: Scenario Aware Autonomous Parking Via\\Integrated Constrained Optimization and Imitation Learning
\vspace{-0.3in}}

\author{Lexiong Huang$^{1,2,*}$, Ruihua Han$^{1,3,*}$, Guoliang Li$^{1}$, He Li$^{4}$, Shuai Wang$^{1,\dagger}$, Yang Wang$^{1,\dagger}$, and Chengzhong Xu$^{4}$
\vspace{-0.2in}
\thanks{This work was supported by the Science and Technology Development Fund of Macao S.A.R (FDCT) (No. 0081/2022/A2), the Guangdong Basic and Applied Basic Research Project (No. 2021B1515120067), the Shenzhen Science and Technology Program (No. RCB20200714114956153), and the Cooperation Project between Shenzhen Institute of Advanced Technology and Direct Drive Tech.}
\thanks{$*$ Equal contribution}
\thanks{$\dagger$ Corresponding author: Shuai Wang (s.wang@siat.ac.cn) and Yang Wang (yang.wang1@siat.ac.cn).}
\thanks{${1}$ Center for Cloud Computing, Shenzhen Institute of Advanced Technology, Chinese Academy of Sciences}
\thanks{${2}$ University of Chinese Academy of Sciences, Beijing 100049, China}
\thanks{${3}$ Department of Computer Science, University of Hong Kong}
\thanks{${4}$ IOTSC, University of Macau}
} 
\vspace{-0.5in}

\begin{document}
\maketitle

\begin{abstract}
Autonomous parking (AP) is an emering technique to navigate an intelligent vehicle to a parking space without any human intervention.
Existing AP methods based on mathematical optimization or machine learning may lead to potential failures due to either excessive execution time or lack of generalization. To fill this gap, this paper proposes an integrated constrained optimization and imitation learning (iCOIL) approach to achieve efficient and reliable AP.
The iCOIL method has two candidate working modes, i.e., CO and IL, and adopts a hybrid scenario analysis (HSA) model to determine the better mode under various scenarios.
We implement and verify iCOIL on the Macao Car Racing Metaverse (MoCAM) platform. 
Results show that iCOIL properly adapts to different scenarios during the entire AP procedure, and achieves significantly larger success rates than other benchmarks.
\end{abstract}

\section{Introduction}\label{section1}

Autonomous parking (AP) is a core task for intelligent driving, which determines a sequence of control commands to move the vehicle from its current position to a parking space without any human intervention \cite{zhang2020optimization, chai2020design,zhang2019reinforcement,du2020trajectory}.
The next-generation AP aims to save more human drivers' parking time by enlarging the AP range from several meters to tens of meters \cite{ahn2022autonomous,han2022differential}.
However, longer AP range requires high adaptation in dynamic scenarios and fast generation of collision-free trajectories.
Optimization-based algorithms \cite{zhang2020optimization,han2022differential} suffer from high computational complexities and may lead to potential failures due to excessive execution time. 
On the other hand, deep learning algorithms \cite{chai2020design,zhang2019reinforcement,du2020trajectory,ahn2022autonomous} can generate driving signals from input images in milliseconds using feed-forward operations.
But they may break down if the target scenario contains data outside the distribution of their training dataset.
Consequently, currently there is no AP algorithm that simultaneously achieves excellent generalization and high computational efficiency. 

This paper considers the integration of optimization and learning methods and aims to realize intelligent switching between the two methods in different scenarios. However, such an integration involves the following technical challenges: 
\begin{itemize}
    \item[(1)] \textbf{Scenario uncertainty for learning algorithms}. The scenario can be complex (i.e., containing static and dynamic objects), heterogeneous (i.e., containing non-IID data with different lighting and weather conditions), and multi-modal (i.e., can be images, point-clouds, or other sensory data) \cite{FLCAV}. 
This makes it difficult to determine whether a given scenario poses a threat to learning algorithms.
    \item[(2)] \textbf{Scenario complexity for optimization algorithms}. 
The computational time of optimization is proportional to the size of state-action space and the number of surrounding objects \cite{canny1988complexity}. This makes it challenging to determine whether a given scenario poses a threat to optimization.
    \item[(3)] \textbf{Real-time implementation of the AP system}. 
This requires highly efficient implementation of the learning and optimization algorithms, and a fast switching mechanism between two algorithms.
\end{itemize}
\begin{figure}[!t]
    \centering
    \begin{subfigure}[t]{0.245\textwidth}
      \includegraphics[height=0.6\textwidth,width=\textwidth]{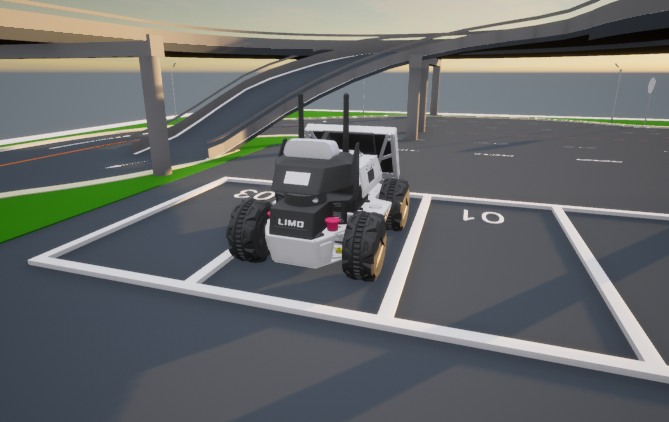}
      \caption{Virtual simulation}
    \end{subfigure}%
    ~
    \begin{subfigure}[t]{0.245\textwidth}
      \includegraphics[height=0.6\textwidth,width=\textwidth]{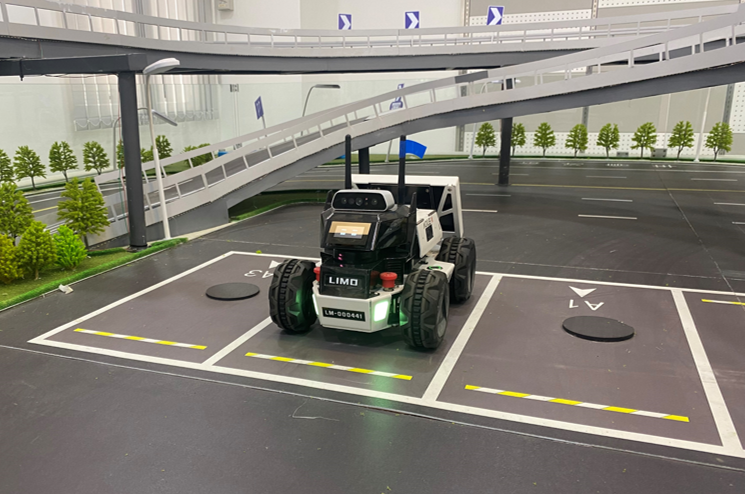}
      \caption{Real-world testbed}
    \end{subfigure}
    \caption{Implementation of iCOIL on MoCAM.}
    \label{fig:IL_compare}
\end{figure}

To address the above challenges, this paper proposes integrated constrained optimization and imitation learning (iCOIL), which serves as an efficient and reliable AP solution. The iCOIL leverages CO to generate collision-free trajectories with theoretical guarantee through mathematical state-evolution models, and IL to mimic human experts' logics and actions through a deep neural network (DNN). To enable transition between the CO and IL modes, iCOIL adopts a hybrid scenario analysis (HSA) model that computes scenario uncertainties based on information entropy and scenario complexities based on geometries of ego-vehicle and obstacles. The iCOIL solution is implemented as a robot operation system (ROS) package and connected to Car Learning to Act (CARLA) \cite{J_Doso17CARLA} via CARLA-ROS bridge. 
As such, the proposed iCOIL-based AP system is validated on the Macao Car Racing Metaverse (MoCAM) platform, which is a digital-twin system that allows design and verification of AP through high-fidelity virtual simulation (i.e, Fig.~1a) and transfers the result to the real-world testbed (i.e., Fig.~1b) through ROS interfaces. 
We also compare the success rate and parking time of iCOIL with IL. 
Experimental results confirm the superiority of iCOIL.
The source codes will be released as open-source ROS packages.

\begin{figure*}[t] 
\centering 
\includegraphics[width=180mm]{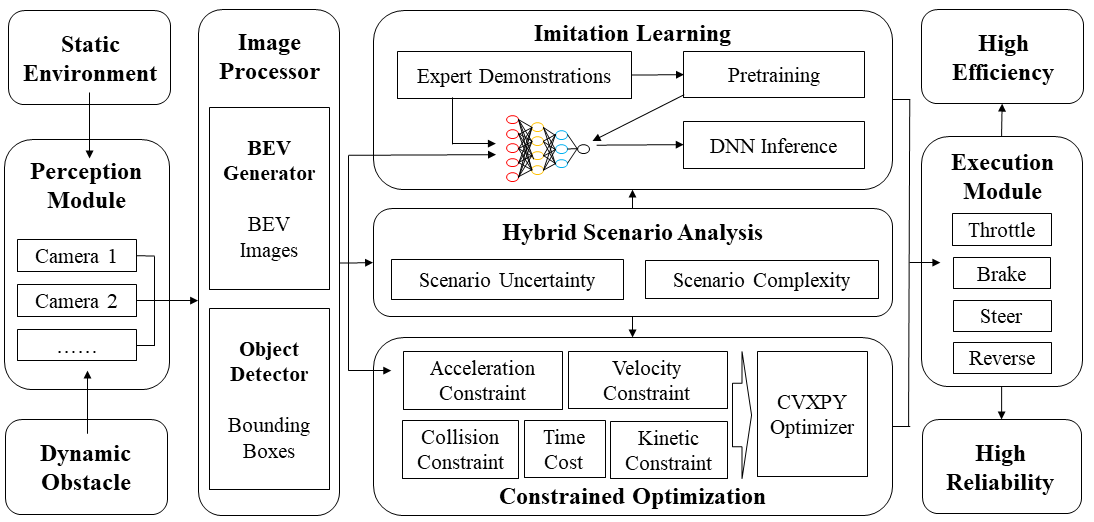}
\caption{The overall architecture of the proposed iCOIL-based AP system.}  
\label{fig:system}  
\vspace{-0.5cm}
\end{figure*}

\section{Related Work}\label{section2} 

\textbf{Learning-based algorithms} can be viewed as black-box inference mappings from input scenarios (i.e., a sequence of images or point-clouds) to driving actions (i.e., throttle, brake, steer, reverse) \cite{chai2020design,zhang2019reinforcement,du2020trajectory,ahn2022autonomous,9351818,hawke2020urban}. 
They do not make assumptions on the sensory and action datasets, but build brain-like DNNs to directly learn from the datasets via back propagation \cite{9351818}. 
Such methods provide efficient feature extractions and representations, as well as fast inference due to non-iterative feed-forward operations. 
Learning-based algorithms can be divided into reinforcement learning (RL) \cite{zhang2019reinforcement,du2020trajectory,9351818,9197209} and IL \cite{chai2020design,ahn2022autonomous,hawke2020urban,tai2016deep,pan2020imitation}. 
The RL algorithm allows the intelligent vehicle to learn from experience by interacting with the environment or other agents \cite{9351818}. 
Nonetheless, the RL-based approaches suffer from long training, work only in confined scenarios, and are sensitive to sensor noises.
IL-based algorithms belong to supervised learning, which mimic the experts' driving policies by learning from demonstration datasets \cite{chai2020design,ahn2022autonomous,hawke2020urban}.  
However, the effectiveness of imitation learning depends on the quality of the dataset and the generalization capability of the DNNs. 
To improve the quality of data, the HG-DAGGER was proposed in \cite{kelly2019hg} to support interactive dataset collection from human experts in real-world systems.
On the other hand, to improve the generalization capability of DNNs, conditional IL \cite{hawke2020urban} and federated IL \cite{liu2021peer} have been proposed for DNN selection and aggregation.

\textbf{Optimization-based algorithms} are hand-designed mathematical operations (mostly iterative) to minimize distances between target and actual waypoints \cite{zhang2020optimization,han2022differential, gonzalez2015review, han2023rda}. 
This can be realized through the model predictive control (MPC) \cite{zhang2020optimization}, which builds the relation between future and current states using a finite horizon Markov decision process. Compared to learning-based algorithms, optimization-based algorithms can generate bounded actions, forcing the vehicle to move within a safety region \cite{han2023rda}. Compared to conventional heuristics such as grid search and random sampling, optimization-based algorithms can achieve higher driving performance since their solution is guaranteed to be local or global optimal \cite{gonzalez2015review}. 
CO-based algorithms are constrained versions of optimization algorithms for cluttered environments with obstacles, which are more challenging to derive than the unconstrained counterparts, since the collision avoidance constraints are nonconvex \cite{zhang2020optimization}. Furthermore, as the number of obstacles increases, these algorithms would require longer execution time than learning or heuristic algorithms \cite{han2023rda}. 
Therefore, recent works on CO-based algorithms mainly focus on how to resolve nonconexity \cite{han2022differential,zhang2020optimization} or accelerate computations \cite{han2023rda}.

\section{The iCOIL-based AP System}\label{section3} 

\begin{figure*}[t] 
    \centering
    \begin{subfigure}[t]{0.247\textwidth}
      \includegraphics[width=\textwidth]{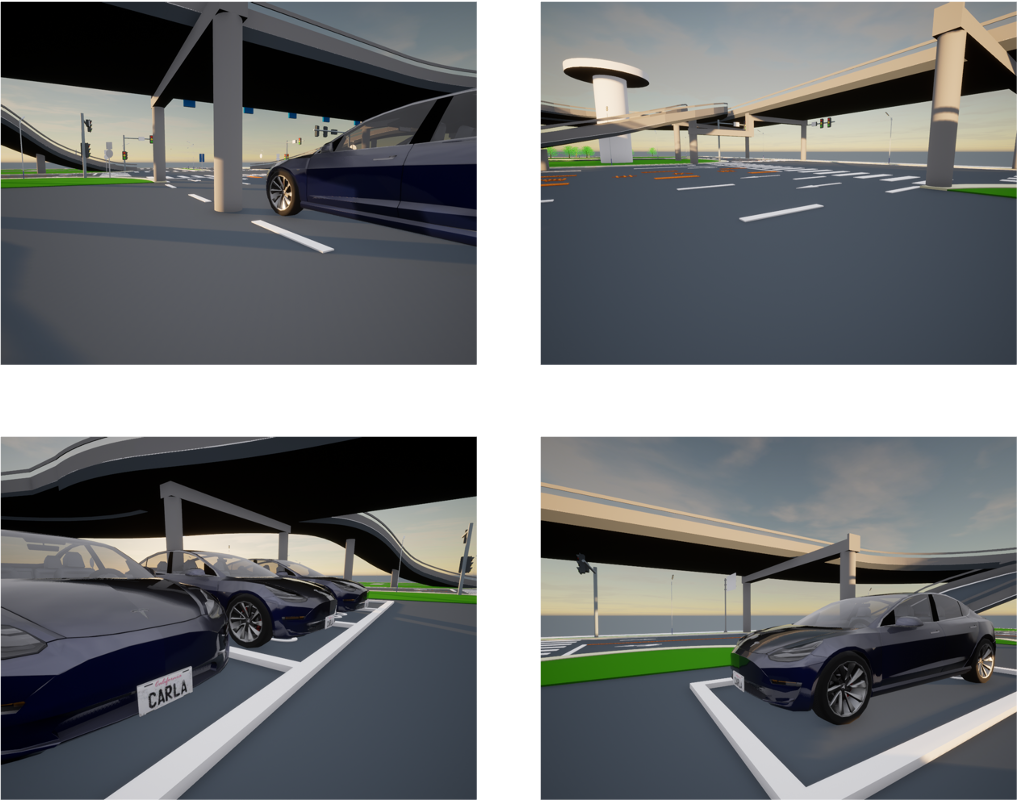}
      \caption{The input ego-view images.}
    \end{subfigure}%
    \quad 
    \begin{subfigure}[t]{0.347\textwidth}
      \includegraphics[width=\textwidth]{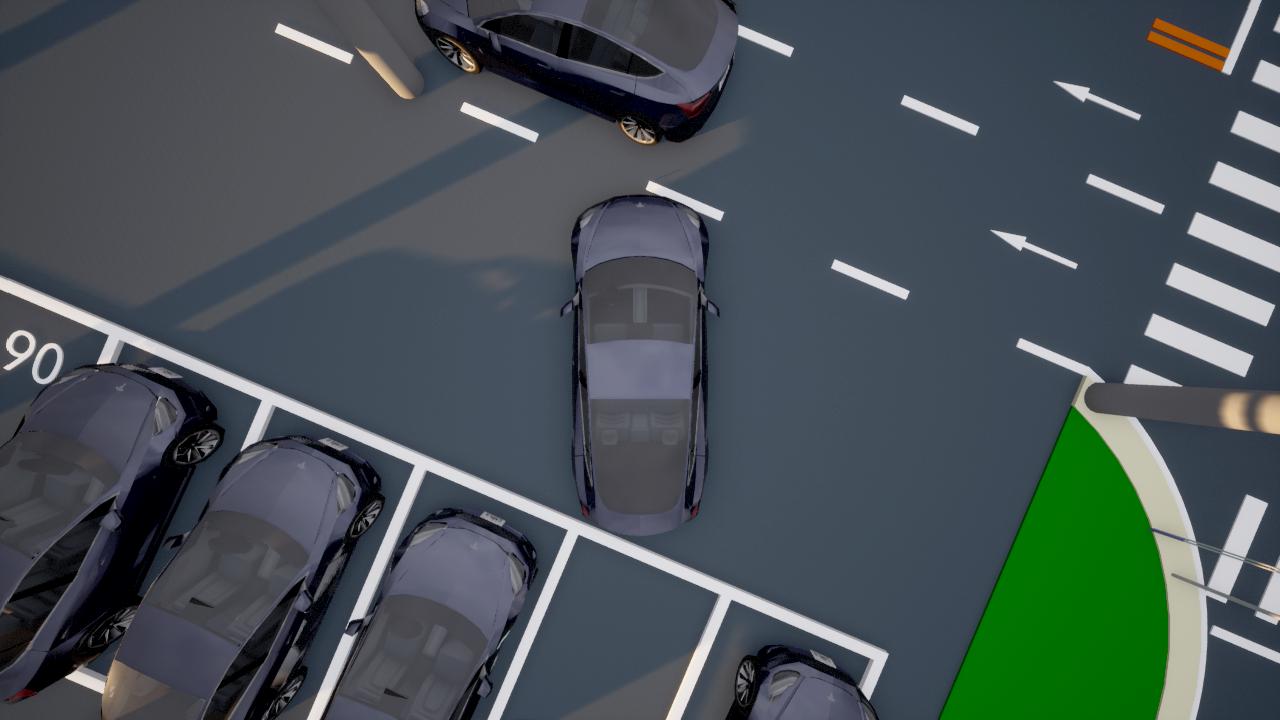}
      \caption{The generated BEV images.}
    \end{subfigure}
    \quad 
    \begin{subfigure}[t]{0.347\textwidth}
      \includegraphics[width=\textwidth]{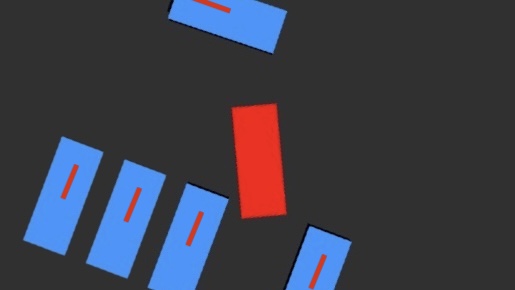}
      \caption{The detected bounding boxes.}
    \end{subfigure}
    \caption{The input data, generated BEV image, and detected boxes for the iCOIL-based AP system.}
    \label{fig:transformer}
\end{figure*}

The \textbf{AP system architecture} is shown in Fig.~\ref{fig:system}, which is written as inference mapping $f: \mathcal{X} \to \mathcal{A}$, where $\mathcal{X}=\{\mathbf{x}_1,\mathbf{x}_2,\cdots\}$ with $\mathbf{x}_i$ being the ego-view images of the $i$-th frame, and $\mathcal{A}=\{\mathbf{a}_1,\mathbf{a}_2,\cdots\}$ with $\mathbf{a}_i$ being the action vector at the $i$-th frame containing throttle, brake, steer, and reverse elements.
At the $i$-th frame, the perception module (i.e., the left hand side of Fig.~\ref{fig:system}) maps environments into images $\mathbf{x}_i$ via onboard cameras. 
The image processor (i.e., next to the perception module) contains a bird's eye view (BEV) transformer $\mathbf{y}_i=g(\mathbf{x}_i)$ and an object detector $\mathbf{z}_i=h(\mathbf{y}_i)$, where $\mathbf{y}_i$ and $\mathbf{z}_i$ are the vectors representing the BEV image and the bounding boxes at the $i$-th frame, respectively.
As shown in Fig.~3, the function $g$ transforms ego-view images (i.e., Fig.~3a) into BEV images (i.e., Fig.~3b), and the function $h$ generates bounding boxes of obstacles (i.e., Fig.~3c).
The proposed iCOIL (i.e., in the middle of Fig.~\ref{fig:system}) contains IL $f_{\mathrm{IL}}(\cdot)$, CO $f_{\mathrm{CO}}, (\cdot)$, and HSA $f_{\mathrm{HSA}}(\cdot)$ modules, which are presented in detail below.

The \textbf{IL module} (i.e., the upper middle part of Fig.~\ref{fig:system}) maps BEV images into actions, i.e., $\mathbf{a}_i=f_{\mathrm{IL}}(\mathbf{y}_i|\bm{\theta}^*)$, where $\bm{\theta}^*$ is the pretrained DNN parameters. 
This $\bm{\theta}^*$ is obtained by collecting the expert demonstrations and training a DNN on that dataset. 
On the other hand, the \textbf{CO module} (i.e., the lower middle part of Fig.~\ref{fig:system}) leverages the detected bounding boxes $\mathbf{z}_i$ to generate collision-free actions,
i.e., $\mathbf{a}_i=f_{\mathrm{CO}}(\mathbf{z}_i)$. 
This $f_{\mathrm{CO}}$ is obtained by solving the AP optimization problem for minimizing a time efficiency cost under a set of constraints on acceleration, velocity, kinetics, and collision avoidance. 

The \textbf{HSA module} (i.e., at the center of Fig.~\ref{fig:system}) 
is a function $f_{\mathrm{HSA}}(\{\mathbf{x}_t\}_{t=i-T}^i)$ that computes the average scenario uncertainty $U_i$ and average scenario complexity $C_i$ at frame $i$ over the past period of time (i.e., a dynamic time window from $t=i-T$ to $t=i$) and uses them as indicators for determining the driving mode. 
If the average uncertainty (complexity) exceeds a certain threshold, then the current scenario poses a threat to the IL (CO) algorithm, and our AP system would switch to the other mode for a higher reliability (efficiency).
Consequently, we can set $f_{\mathrm{HSA}}(\{\mathbf{x}_t\}_{t=i-T}^i)=U_iC_i^{-1}$ and adopt the following conditions for mode switching:
\begin{equation}\label{ms}
f(\mathbf{x}_i)=
\left\{\begin{array}{ll}
f_{\mathrm{IL}}\left(g(\mathbf{x}_i)|\bm{\theta}^*\right), & \mathrm{if}~f_{\mathrm{HSA}}(\{\mathbf{x}_t\}_{t=i-T}^i)\leq \lambda\\
f_{\mathrm{CO}}\left(h(g(\mathbf{x}_i))\right), &\mathrm{if}~f_{\mathrm{HSA}}(\{\mathbf{x}_t\}_{t=i-T}^i)>\lambda
\end{array}\right.
,
\end{equation} 
where $\lambda$ is a predefined threshold to be fine-tuned empirically. 

\section{The iCOIL Algorithm Design}\label{section4} 

\subsection{IL}

For the proposed IL module, the continuous driving actions are converted to discrete values, so as to formulate IL as a multi-category classification problem \cite{chai2020design,ahn2022autonomous,hawke2020urban}. 
This make it possible to leverage a DNN consisting of a feature extraction network and a state-action network, where the former extracts high dimensional features from images and the latter maps features into actions. 
The feature extraction network adopts three layers, and each layer has three units, i.e., convolution, ReLU activation, and max pooling. 
The state-action network has four fully connected layers followed by a soft-max layer that outputs probabilistic actions. 
The action with the highest probability is selected for actual execution at the ego-vehicle. All parameters of the DNN are collectively denoted as $\bm{\theta}$.

Based on the above structure, the DNN training is an optimization problem w.r.t. $\bm{\theta}$, with the objective of minimizing the difference between the DNN outputs and the expert actions:
\begin{equation}
    \bm{\theta}^*=\arg\min\limits_{\bm{\theta}}~
    \sum_{(\mathbf{x}_i',\mathbf{a}_i')\in \mathcal{D}}L\left(f_{\mathrm{IL}}\left(g(\mathbf{x}_i')|\bm{\theta}\right), \mathbf{a}_i' \right),
\end{equation}
where $\mathcal{D}$ denotes the training dataset, and $L(\cdot)$ is the cross entropy function
\begin{equation}
    L=-\frac{1}{|\mathcal{D}|}\sum^M_j A_{i,j}'\log(f_{{\mathrm{IL},j}}^{\mathrm{Prob}}\left(g(\mathbf{x}_i')|\bm{\theta}\right))
\end{equation}
where $|\mathcal{D}|$ is the cardinality of $\mathcal{D}$ (i.e., the number of samples), $M$ is the number of classes after action discretization, $A_{i,j}'$ is the $j$-th element of $\mathbf{a}_i'$, and 
and $f_{\mathrm{IL},j}^{\mathrm{Prob}}(\cdot)$ denotes the probabilistic value of the $j$-th DNN output.

\subsection{CO}

The CO module leverages the Ackermann kinetics model $\mathbf{s}_{i+1}=u(\mathbf{s}_{i},\mathbf{a}_i)$ to estimate the trajectories $\mathbf{s}_{i+1},\cdots,\mathbf{s}_{i+H}$, where 
$\mathbf{s}_{i}$ denotes the location and orientation of ego-vehicle at frame $i$ obtained from $h(g(\mathbf{x}_i))$, and $H$ is the length of prediction horizon.
With the above kinetics model, 
we can derive function $\mathbf{s}_{i+h+1}=v_{h+1}(\mathbf{s}_{i},\{\mathbf{a}_t\}_{t=i}^{i+h})$ for all $h\in[0,H-1]$, where $v_1=u$.
The goal of AP is to 
move the vehicle into the parking space using the minimum time. 
This can be realized via minimizing the distance cost \cite{zhang2020optimization}
\begin{equation}
    C(\{\mathbf{a}_t\}_{t=i}^{i+H-1})=\sum^{H-1}_{h=0} 
    \left\|v_{h+1}(\mathbf{s}_{i},\{\mathbf{a}_t\}_{t=i}^{i+h})-\mathbf{s}_{i+h+1}^*\right\|^2,
\end{equation}
where $\{\mathbf{s}_{i}^*\}$ represent the list of target waypoints, e.g., the shortest path from the current position to the target parking space. 
On the other hand, the ego-vehicle is not allowed to collide with any surrounding objects:
\begin{equation}
    \left\|v_{h+1}(\mathbf{s}_{i},\{\mathbf{a}_t\}_{t=i}^{i+h})-\mathbf{o}_{h+1,k}\right\| \geq d_{\mathrm{safe},k}, \forall h,k
\end{equation}
where $\mathbf{o}_{h,k}$ is the position of the $k$-th obstacle at frame $i+h$ and $d_{\mathrm{safe},k}$ is the safety distance related to the size of obstacle $k$. 
Based on the above analysis, the collision-free AP is formulated as the following CO problem \cite{han2023rda}: 
\begin{equation}\label{MPC}
    \{\mathbf{a}_t^*\}_{t=i}^{i+H-1}=\mathop{\arg\min}_{\mathbf{a}_t\in\mathcal{A},\forall t}
    \left\{C(\{\mathbf{a}_t\}_{t=i}^{i+H-1}): (5)\right\},
\end{equation}
where $\mathcal{A}$ is the boundary set for driving actions. 
The above problem is nonconvex due to the functions $\{v_{h+1}(\cdot)\}$ in both objective and constraints.
We convert the primal problem into several convex optimization problems and solve them using the open-source optimization software (e.g., CVXPY). 
The output action at frame $i$ is set to $f_{\mathrm{CO}}\left(h(g(\mathbf{x}_i))\right)=\mathbf{a}_i^*$.

\subsection{HSA}

The HSA module needs to compute scenario uncertainty $U_i$ and scenario complexity $C_i$ over the past $T$ frames. 
In particular, the scenario uncertainty measures the ``confidence'' of IL, and is related to the probabilistic distribution of the DNN outputs \cite{kendall2017uncertainties,kendall2018multi}. 
For instance, even distributions indicate larger uncertainties while non-even output distributions indicate smaller uncertainties \cite{kendall2017uncertainties,kendall2018multi}.
Consequently, the instant scenario uncertainty $\omega_i$ at frame $i$ is defined as the entropy of the probabilistic distribution of the DNN outputs $\omega_i=-\sum^M_{j=1}p_{i,j}\,\log{p_{i,j}}$, where $p_{i,j}$ is the probability value for action $j$ at frame $i$.
By summing up $\omega_i$ over the past $T$ frames, we have 
\begin{align}
U_i=\frac{1}{T}\sum_{h=0}^{T-1}\omega_{i-h}=
-\frac{1}{T}\sum_{h=0}^{T-1}\sum^M_{j=1}p_{i-h,j}\,\log{p_{i-h,j}}
.
\end{align}

On the other hand, the scenario complexity measures the computational delay of CO, which is proportional to the computational complexity for solving \eqref{MPC}. 
By averaging the computational complexity over the past $T$ frames, the average scenario complexity is \cite{han2023rda}
\begin{align}\label{Ci}
    C_i=&\frac{1}{T}\sum^{T-1}_{h=0}\left[H\left(N_a+\sum^K_{k=1}  \mathrm{e}^{-\|D_0-D_{i-h,k}\|}\right)\right]^{3.5}.
\end{align}
where the parameters $H, N_a, D_0, D_{j,k}$ are defined in Table I.

\textbf{Interpretation of $C_i$}: 
First, the power of $3.5$ in equation \eqref{Ci} indicates that the computational delay increases superlinearly with the length of prediction horizon and the number of obstacles.
Second, the term $\mathrm{e}^{-\|D_0-D_{j,k}\|}$ in equation \eqref{Ci} means that only a part of obstacles may have impact on the computational complexity, where $D_0$ is the most dangerous obstacle position.
This is because the planning space is reduced if the obstacle is very close, and the collision probability is close to zero if the obstacle is far-away \cite{canny1988complexity}. 

\begin{table}[!t]
    \centering
    \caption{Parameters for scenario complexity}
    \begin{tabular}{c|c|c}
        \hline
        Type & Symbol & Parameters\\
        \hline
        Ego-vehicle & $H$ & length of prediction horizon \\
        Ego-vehicle & $N_a$ & dimension of action space\\
        \hline
        Obstacle & $D_0$ & most dangerous obstacle distance \\
        Obstacle & $D_{j,k}$ & distance of obstacle $k$ at frame $j$\\
        \hline
    \end{tabular}
\end{table}

\begin{figure}
    \centering
    \includegraphics[width=0.45\textwidth]{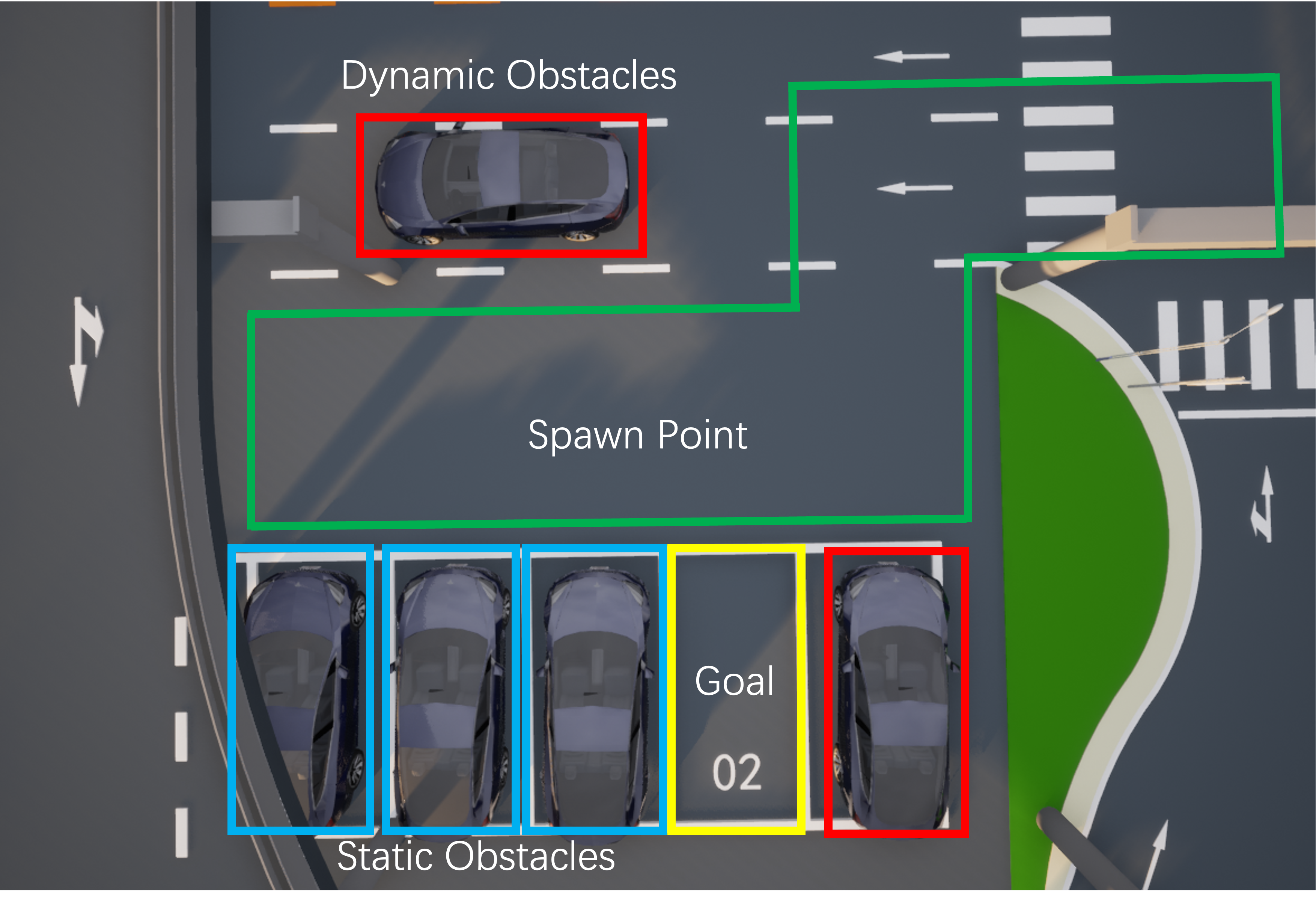}
    \caption{The simulated AP scenario in CARLA.}
    \label{fig:test}
\end{figure}

\begin{figure}[!t]
    \centering
    \begin{subfigure}[t]{0.245\textwidth}
      \includegraphics[width=\textwidth]{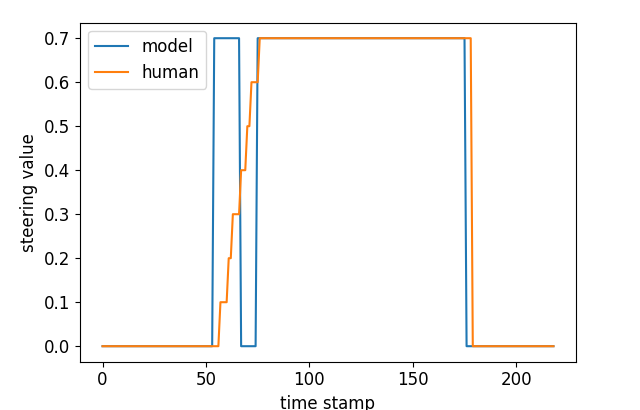}
      \caption{}
    \end{subfigure}%
    ~
    \begin{subfigure}[t]{0.245\textwidth}
      \includegraphics[width=\textwidth]{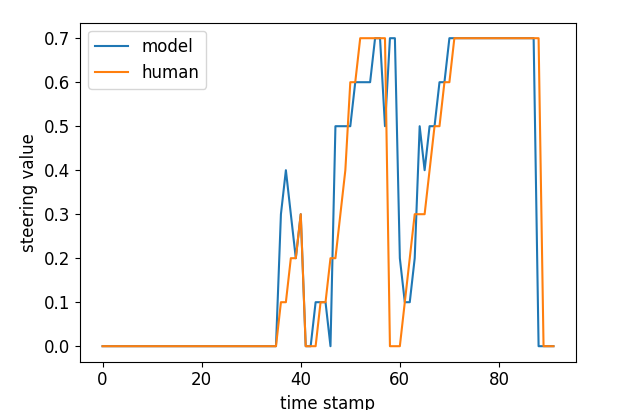}
      \caption{}
    \end{subfigure}
    \caption{Steering values of IL and human driver.}
    \label{fig:IL_compare}
\end{figure}

\section{Experiments}\label{section5}

\begin{figure*}[t] 
    \centering
    \begin{subfigure}[t]{0.23\textwidth}
      \includegraphics[width=\textwidth]{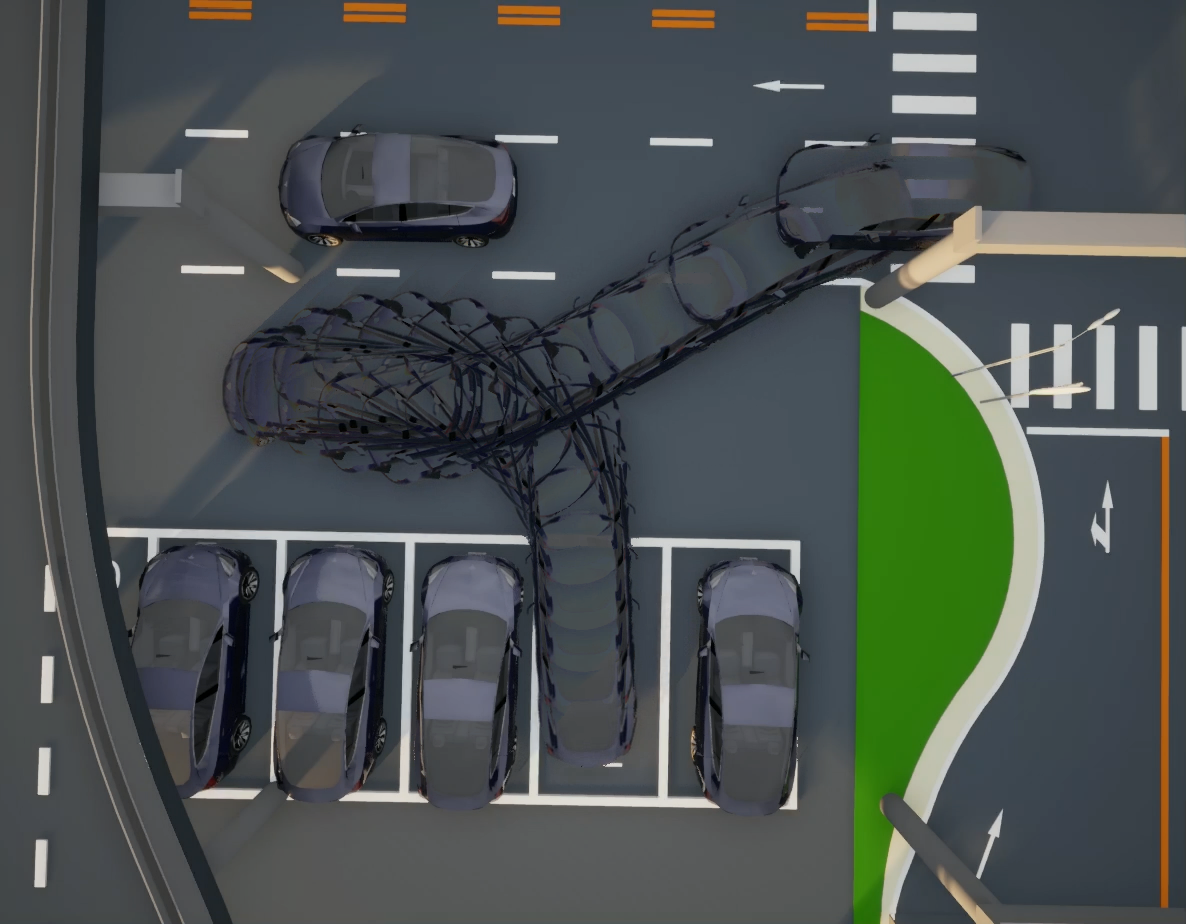}
      \caption{}
    \end{subfigure}%
    \quad 
    \begin{subfigure}[t]{0.23\textwidth}
      \includegraphics[width=\textwidth]{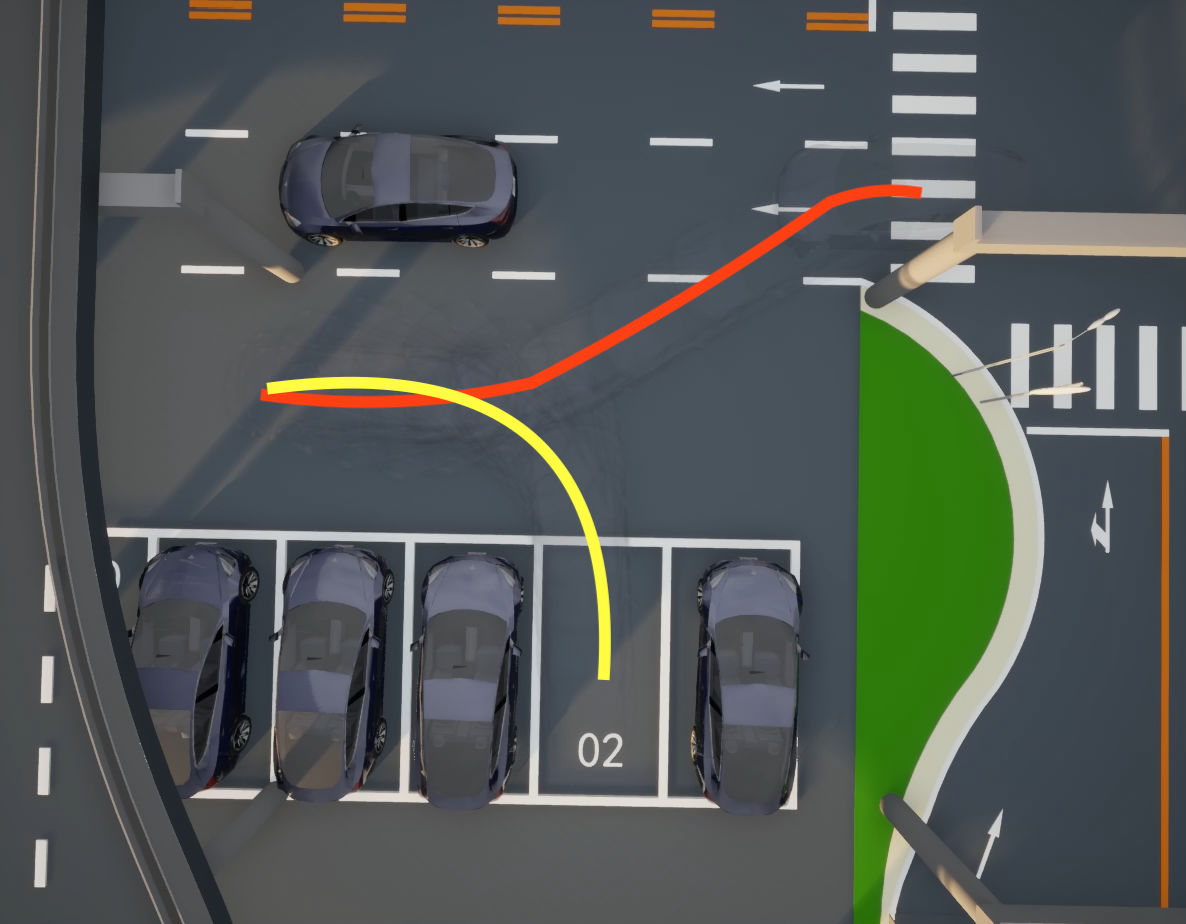}
        \caption{}
    \end{subfigure}
    \quad 
    \begin{subfigure}[t]{0.23\textwidth}
      \includegraphics[width=\textwidth]{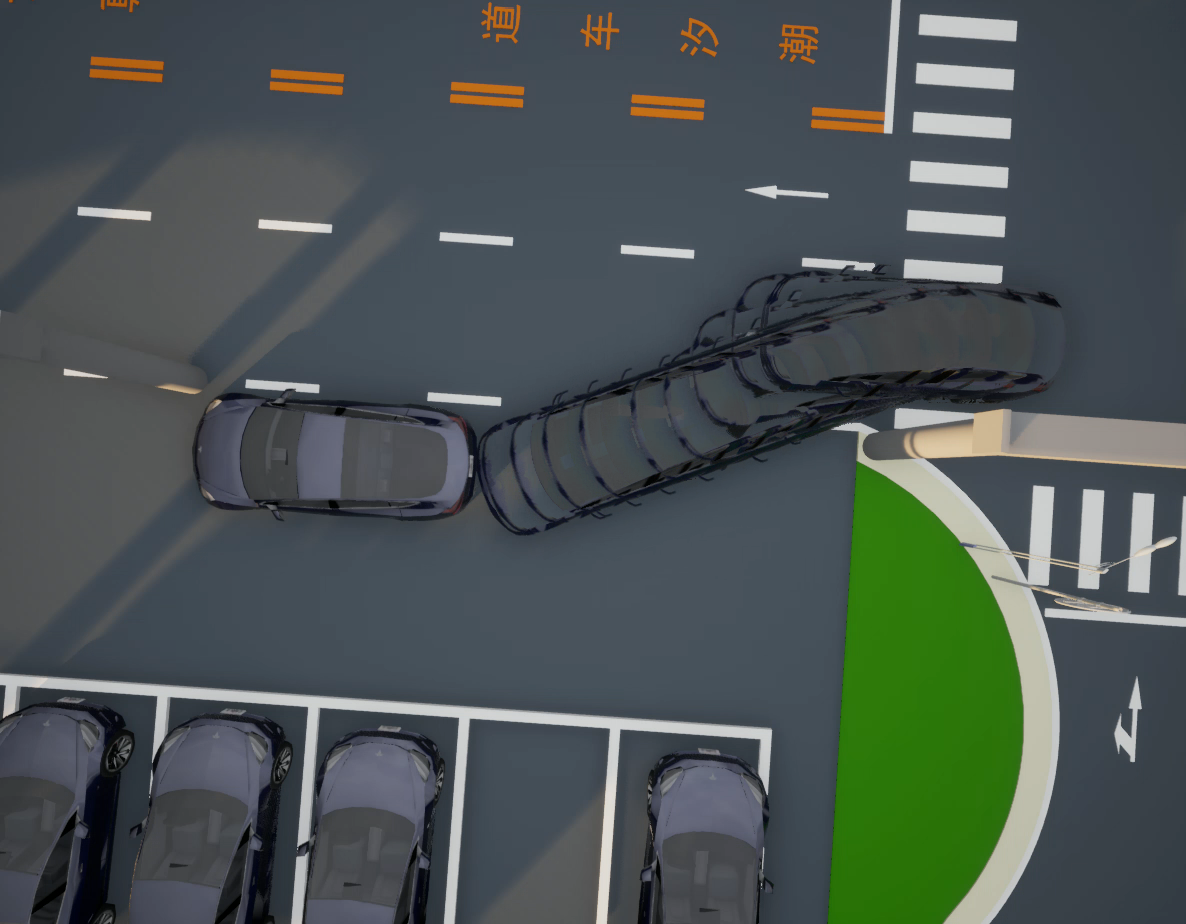}
        \caption{}
    \end{subfigure}
    \quad 
    \begin{subfigure}[t]{0.23\textwidth}
      \includegraphics[width=\textwidth]{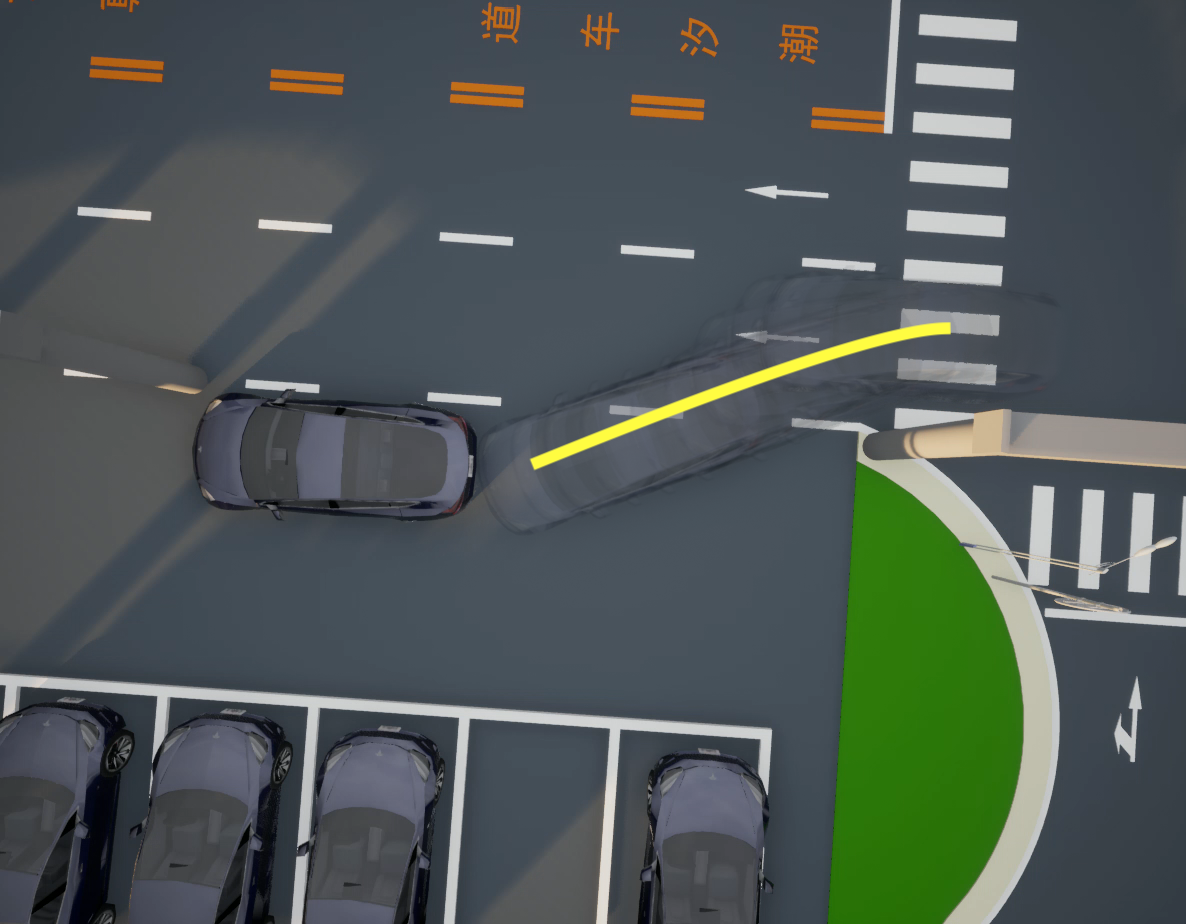}
        \caption{}
    \end{subfigure}
    \caption{Comparison of the parking processes and trajectories between iCOIL and IL. (a) The parking process of the iCOIL-based AP. (b) The trajectory of the iCOIL-based AP. (c) The parking process of the IL-based AP. (d) The trajectory of the IL-based AP. The red curve represents the CO mode and the yellow curve represents the IL mode.     
    }
    \label{fig:trajectory}
\end{figure*}

\begin{figure}[!ht]
  \centering
  \subfloat{\includegraphics[width=0.52\textwidth]{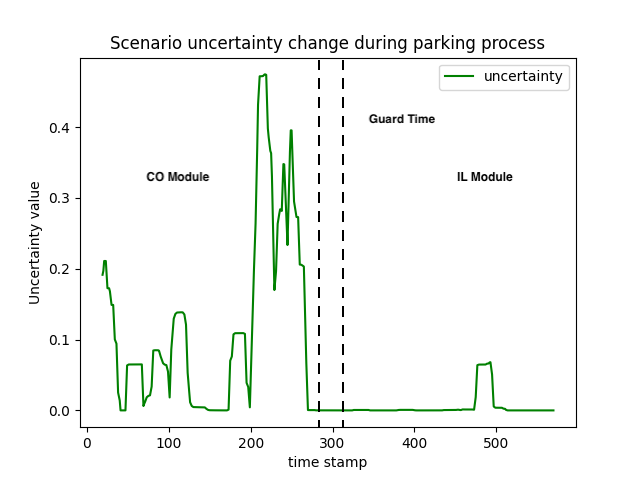}}\hspace{20pt}
  \subfloat{\includegraphics[width=0.52\textwidth]{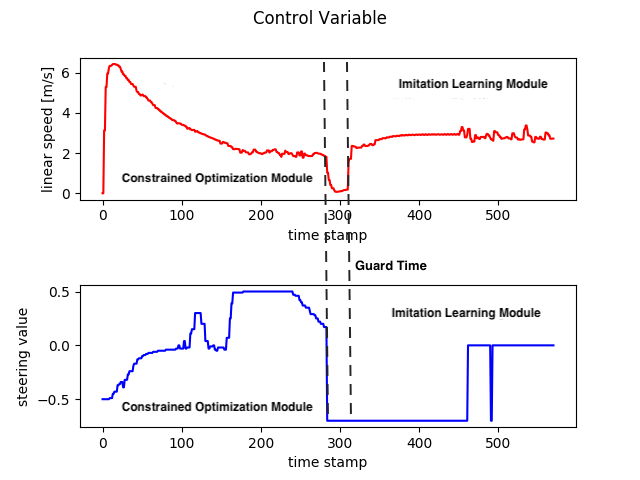}}
  \caption{Mode switching based on HSA.}
  \label{fig:anomaly}
\end{figure}

\subsection{Implementation Details}

The Macao Car Racing Metaverse (MoCAM) platform is adopted to verify the performance of the proposed iCOIL-based AP. 
MoCAM is a digital-twin system consisting of a real-world sandbox and a high-fidelity virtual simulator developed based on CARLA \cite{J_Doso17CARLA}. 
The iCOIL is implemented as three Python ROS nodes (i.e., IL, CO, and HSA nodes), and connected to MoCAM via CARLA-ROS bridge.
Besides MoCAM and iCOIL nodes, we also deploy BEV transformer and object detector nodes, which are off-the-shelf open-source software. 
Data sharing among all these registered nodes is realized via ROS communications, where the nodes publish or subscribe ROS topics that carry the information of ego-view images, BEV images, bounding boxes, and control commands.
Besides the proposed iCOIL, we also realize the conventional IL scheme \cite{chai2020design} for comparison, which directly adopts a multi-layer DNN for AP. 
All experiments are conducted on a desktop with Intel i9 CPU and NVIDIA 3080 GPU.

\subsection{Settings}

We consider the map shown in Fig.~\ref{fig:test}, where the starting point of ego-vehicle is randomly generated within the spawn point region (i.e., green area), and the parking space is within the goal region (i.e., yellow box). The AP task can be classified into: easy level (i.e., there are only three static obstacles marked in blue), normal level (i.e., there are three static obstacles and two dynamic obstacles marked in red), and hard level (i.e., all obstacles exist and we add additional noises to the input images and bounding boxes).
The adversarial data can simulate the real-world uncertainties and verify the robustness of the proposed iCOIL.

\subsection{Validation of the Proposed iCOIL-based AP}

We adopt MoCAM to generate the driving dataset for training the IL DNN. 
We collect $5171$ data samples ($2624$ for forward-moving and $2547$ for reverse-parking) and terminate the training process after 300 epochs. 
The steering actions generated by the IL and the human driver are compared in Fig.~\ref{fig:IL_compare}. 
It can be seen that the trained IL generates similar actions to those of the human driver.
However, since discretization is applied to the action space, the action curve of IL is stepped and less smooth than that of the human driver.

To verify the effectiveness of the iCOIL-based AP, a complete parking process from the starting to end points is shown in Fig.~6a. 
It can be seen that the proposed iCOIL-based AP successfully navigates the ego-vehicle to the target parking space without any collision.
The scenario uncertainty generated by the HSA module during the above procedure is shown in Fig.~\ref{fig:anomaly}.
The uncertainty fluctuates between $0$ to $0.5$ at the beginning, but after around $280$ time stamps, it drops to below $0.1$ and keeps stable. 
This demonstrates the excellent scene understanding capability of the proposed HSA.
Based on HSA, the iCOIL-based AP system performs mode switching from CO to IL for improving the AP efficiency as shown in Fig.~6b. 
The associated control commands over time are shown in Fig.~\ref{fig:anomaly}.
The reverse button is turned on after mode switching and the steering value is close to zero after around $470$ time stamps, meaning that the ego-vehicle enters the parking space and moves backwards slowly to complete the AP task. 
Note that a guard time with $20$ time stamps is added in our implementation to smooth the transition between different modes. 
The above experiment demonstrates the effectiveness of the proposed iCOIL-based AP.

\subsection{Performance Comparison}

This subsection compares the parking time and success rates of the iCOIL and IL schemes. 
The parking time is defined as the total amount of time from the starting point to the parking space. 
The task is deemed as failed if the ego-vehicle cannot reach the goal within a given time or collides with other obstacles; otherwise successful.

The parking time and success rates under different levels of difficulty are shown in Table~\ref{tab:compare}.
For the easy level, iCOIL achieves $94\%$ success rate and the IL achieves $72\%$ success rate. But the parking time of IL is slightly shorter. 
For the normal level, due to the consideration of dynamic obstacles, the success rate declined for both methods. However, the success rate of the IL drops significantly due to insufficient generalization. 
A snapshot of parking failure with IL is shown in Fig.~6c and 6d. 
In contrast, the success rate of iCOIL is not much affected. 
This is because the iCOIL scheme adopts CO for the collision avoidance, which generates more accurate and safe actions than those of IL.
As for the hard level, the performance of IL is further degraded, since the sensing uncertainty is enlarged. 
In all difficulty levels, the proposed iCOIL always outperformed the IL in terms of success rate metrics.
Note that the success rate of IL could be enhanced with more data and larger DNNs \cite{chai2020design}. 
In this case, the gap between IL and iCOIL could be smaller, but the iCOIL would still outperform IL in corner case scenarios.

\begin{table}[!t]
    \centering
    \caption{Comparison of parking time and success rate}
    \label{tab:compare}
    \begin{tabular}{ccccc}
        \hline
        \multicolumn{5}{c}{Easy Task} \\
        \hline
        Method & Average & Max & Min & Success Ratio\\
        \hline
        iCOIL & 26.02 & 27.21 & 24.89 & 94\% \\
        IL \cite{chai2020design} & 23.65 & 25.16 & 22.52 & 72\% \\
        \hline
    \end{tabular}
    \hspace{0.5in}
    \begin{tabular}{ccccc}
        \hline
        \multicolumn{5}{c}{Normal Task} \\
        \hline
        Method & Average & Max & Min & Success Ratio\\
        \hline
        iCOIL & 25.40 & 26.29 & 24.01 & 91\% \\
        IL \cite{chai2020design} & 25.81 & 26.54 & 23.77 & 36\% \\
        \hline
    \end{tabular}
    \hspace{0.5in}
    \begin{tabular}{ccccc}
        \hline
        \multicolumn{5}{c}{Hard Task} \\
        \hline
        Method & Average & Max & Min & Success Ratio\\
        \hline
        iCOIL & 25.72 & 26.70 & 24.58 & 92\% \\
        IL \cite{chai2020design} & 24.12 & 26.44 & 23.31 & 33\% \\
        \hline
    \end{tabular}
\end{table}

\subsection{Sensitivity Analysis}

Finally, we validate the parking performance of iCOIL at different starting points. 
In particular, we consider close, remote, and random starting points in Fig.~~\ref{fig:test} and the associated results under different numbers of obstacles are shown in Fig.~8. 
For the close starting point case, the number of obstacles has little impact on the parking time. 
However, as for remote and random cases, the parking time increases as the number of obstacles increases.
This is because the starting point may be close to the obstacles in such a case, which significantly increases the parking time.
Note that for the case with random starting points, the fluctuation of parking time is large, as the parking distance may vary significantly under different moving trajectories. 

\begin{figure}[!t]
    \centering
\label{fig:res2}
      \includegraphics[width=0.49\textwidth]{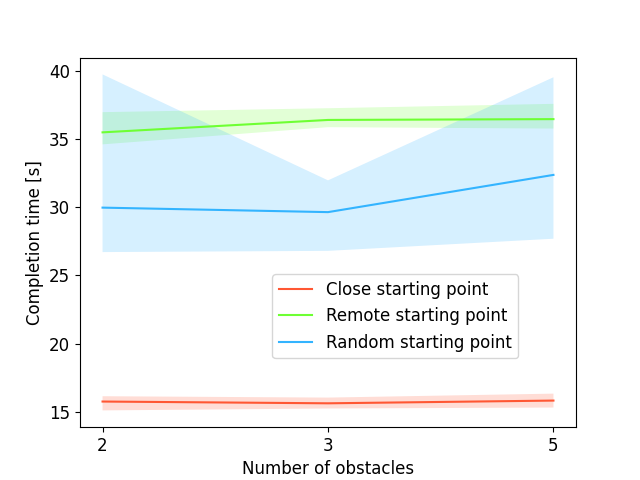}
      \caption{The parking time of iCOIL under different starting points and numbers of obstacles.}
    \caption{Comparison of parking time.}
\end{figure}

Lastly, we test the execution frequency of iCOIL. 
The average frequencies of IL and CO are 75Hz and 18Hz, respectively. 
This corroborates our theoretic analysis and demonstrates the necessity of mode switching. 
Note that both frequencies can meet the requirement of low-speed parking scenarios.

\section{Conclusion}\label{section6}

This paper presented iCOIL, a scenario-aware AP solution for achieving high efficiency and reliability by integrating CO and IL.
The iCOIL was implemented as a ROS package and deployed in a real-time AP system. 
We validated the iCOIL-based AP system in MoCAM and demonstrated its effectiveness in various configurations. 
The iCOIL scheme was shown to outperform the conventional IL in terms of the success rate.

\bibliographystyle{IEEEtran}
\bibliography{reference/Thesis}

\begin{thebibliography}{10}
\providecommand{\url}[1]{#1}
\csname url@samestyle\endcsname
\providecommand{\newblock}{\relax}
\providecommand{\bibinfo}[2]{#2}
\providecommand{\BIBentrySTDinterwordspacing}{\spaceskip=0pt\relax}
\providecommand{\BIBentryALTinterwordstretchfactor}{4}
\providecommand{\BIBentryALTinterwordspacing}{\spaceskip=\fontdimen2\font plus
\BIBentryALTinterwordstretchfactor\fontdimen3\font minus
  \fontdimen4\font\relax}
\providecommand{\BIBforeignlanguage}[2]{{%
\expandafter\ifx\csname l@#1\endcsname\relax
\typeout{** WARNING: IEEEtran.bst: No hyphenation pattern has been}%
\typeout{** loaded for the language `#1'. Using the pattern for}%
\typeout{** the default language instead.}%
\else
\language=\csname l@#1\endcsname
\fi
#2}}
\providecommand{\BIBdecl}{\relax}
\BIBdecl

\bibitem{zhang2020optimization}
X.~Zhang, A.~Liniger, and F.~Borrelli, ``Optimization-based collision
  avoidance,'' \emph{IEEE Transactions on Control Systems Technology}, vol.~29,
  no.~3, pp. 972--983, May. 2020.

\bibitem{chai2020design}
R.~Chai, A.~Tsourdos, A.~Savvaris, S.~Chai, Y.~Xia, and C.~P. Chen, ``Design
  and implementation of deep neural network-based control for automatic parking
  maneuver process,'' \emph{IEEE Transactions on Neural Networks and Learning
  Systems}, vol.~33, no.~4, pp. 1400--1413, April. 2022.

\bibitem{zhang2019reinforcement}
P.~Zhang, L.~Xiong, Z.~Yu, P.~Fang, S.~Yan, J.~Yao, and Y.~Zhou,
  ``Reinforcement learning-based end-to-end parking for automatic parking
  system,'' \emph{Sensors}, vol.~19, no.~18, p. 3996, September. 2019.

\bibitem{du2020trajectory}
Z.~Du, Q.~Miao, and C.~Zong, ``Trajectory planning for automated parking
  systems using deep reinforcement learning,'' \emph{International Journal of
  Automotive Technology}, vol.~21, pp. 881--887, July. 2020.

\bibitem{ahn2022autonomous}
J.~Ahn, M.~Kim, and J.~Park, ``Autonomous driving using imitation learning with
  look ahead point for semi structured environments,'' \emph{Scientific
  Reports}, vol.~12, no.~1, p. 21285, December. 2022.

\bibitem{han2022differential}
Z.~Han, Y.~Wu, T.~Li, L.~Zhang, L.~Pei, L.~Xu, C.~Li, C.~Ma, C.~Xu, S.~Shen
  \emph{et~al.}, ``Differential flatness-based trajectory planning for
  autonomous vehicles,'' \emph{arXiv preprint arXiv:2208.13160}, 2022.

\bibitem{FLCAV}
S.~Wang, C.~Li, D.~W.~K. Ng, Y.~C. Eldar, H.~V. Poor, Q.~Hao, and C.~Xu,
  ``Federated deep learning meets autonomous vehicle perception: Design and
  verification,'' \emph{IEEE Network}, pp. 1--10, December. 2022.

\bibitem{canny1988complexity}
J.~Canny, \emph{The complexity of robot motion planning}.\hskip 1em plus 0.5em
  minus 0.4em\relax MIT press, 1988.

\bibitem{J_Doso17CARLA}
A.~Dosovitskiy, G.~Ros, F.~Codevilla, A.~Lopez, and V.~Koltun, ``{CARLA}: An
  open urban driving simulator,'' in \emph{Proc. CoRL}, 2017, pp. 1--16.

\bibitem{9351818}
B.~R. Kiran, I.~Sobh, V.~Talpaert, P.~Mannion, A.~A.~A. Sallab, S.~Yogamani,
  and P.~Pérez, ``Deep reinforcement learning for autonomous driving: A
  survey,'' \emph{IEEE Transactions on Intelligent Transportation Systems},
  vol.~23, no.~6, pp. 4909--4926, June. 2022.

\bibitem{hawke2020urban}
J.~Hawke, R.~Shen, C.~Gurau, S.~Sharma, D.~Reda, N.~Nikolov, P.~Mazur,
  S.~Micklethwaite, N.~Griffiths, A.~Shah \emph{et~al.}, ``Urban driving with
  conditional imitation learning,'' in \emph{2020 IEEE International Conference
  on Robotics and Automation (ICRA)}, Paris, France, 2020, pp. 251--257.

\bibitem{9197209}
R.~Han, S.~Chen, and Q.~Hao, ``Cooperative multi-robot navigation in dynamic
  environment with deep reinforcement learning,'' in \emph{2020 IEEE
  International Conference on Robotics and Automation (ICRA)}, Paris, France,
  2020, pp. 448--454.

\bibitem{tai2016deep}
L.~Tai, S.~Li, and M.~Liu, ``A deep-network solution towards model-less
  obstacle avoidance,'' in \emph{2016 IEEE/RSJ international conference on
  intelligent robots and systems (IROS)}, Daejeon, Korea (South), 2016, pp.
  2759--2764.

\bibitem{pan2020imitation}
Y.~Pan, C.-A. Cheng, K.~Saigol, K.~Lee, X.~Yan, E.~A. Theodorou, and B.~Boots,
  ``Imitation learning for agile autonomous driving,'' \emph{The International
  Journal of Robotics Research}, vol.~39, no. 2-3, pp. 286--302, October. 2019.

\bibitem{kelly2019hg}
M.~Kelly, C.~Sidrane, K.~Driggs-Campbell, and M.~J. Kochenderfer, ``Hg-dagger:
  Interactive imitation learning with human experts,'' in \emph{2019
  International Conference on Robotics and Automation (ICRA)}, Montreal, QC,
  Canada, 2019, pp. 8077--8083.

\bibitem{liu2021peer}
B.~Liu, L.~Wang, X.~Chen, L.~Huang, D.~Han, and C.-Z. Xu, ``Peer-assisted
  robotic learning: a data-driven collaborative learning approach for cloud
  robotic systems,'' in \emph{2021 IEEE International Conference on Robotics
  and Automation (ICRA)}, Xi'an, China, 2021, pp. 4062--4070.

\bibitem{gonzalez2015review}
D.~Gonz{\'a}lez, J.~P{\'e}rez, V.~Milan{\'e}s, and F.~Nashashibi, ``A review of
  motion planning techniques for automated vehicles,'' \emph{IEEE Transactions
  on intelligent transportation systems}, vol.~17, no.~4, pp. 1135--1145,
  April. 2016.

\bibitem{han2023rda}
R.~Han, S.~Wang, S.~Wang, Z.~Zhang, Q.~Zhang, Y.~C. Eldar, Q.~Hao, and J.~Pan,
  ``Rda: An accelerated collision free motion planner for autonomous navigation
  in cluttered environments,'' \emph{IEEE Robotics and Automation Letters},
  vol.~8, no.~3, pp. 1715--1722, March. 2023.

\bibitem{kendall2017uncertainties}
A.~Kendall and Y.~Gal, ``What uncertainties do we need in bayesian deep
  learning for computer vision?'' in \emph{Advances in Neural Information
  Processing Systems}, vol.~30, Long Beach, CA, USA, 2017.

\bibitem{kendall2018multi}
A.~Kendall, Y.~Gal, and R.~Cipolla, ``Multi-task learning using uncertainty to
  weigh losses for scene geometry and semantics,'' in \emph{Proceedings of the
  IEEE conference on computer vision and pattern recognition}, Salt Lake City,
  UT, USA, 2018, pp. 7482--7491.

\end{thebibliography}

\end{document}